\documentclass[12pt]{article}

\usepackage{fullpage}

\title{Understanding AI Data Repositories with Automatic Query Generation}
\author{Erik Altman}
\date{}

\begin{document}
\maketitle

\begin{abstract}

We describe a set of techniques to generate queries automatically based on
one or more ingested, input corpuses.  These queries require no {\it a
priori} domain knowledge, and hence no human domain experts.  Thus, these
auto-generated queries help address the epistemological question of how we
know what we know, or more precisely in this case, how an AI system with
ingested data knows what it knows.  

These auto-generated queries can also be used to identify and remedy problem
areas in ingested material -- areas for which the knowledge of the AI system
is incomplete or even erroneous.  Similarly, the proposed techniques
facilitate tests of AI capability -- both in terms of coverage and accuracy.

By removing humans from the main learning loop, our approach also allows more
effective scaling of AI and cognitive capabilities to provide (1) broader
coverage in a single domain such as health or geology; and (2) more rapid
deployment to new domains.  The proposed techniques also allow ingested
knowledge to be extended naturally.

Our investigations are early, and this paper provides a description of the
techniques.  Assessment of their efficacy is our next step for future work.

\end{abstract}

\def\baselinestretch{0.90}

\section{Introduction}
\label{section-Intro}

Broadly speaking, AI or cognitive solutions must follow the {\it ``URL''}
principles:  {\it Understand, Reason, Learn}.  A core challenge for all three
of these URL aspects is ingesting input material on specific topics, and then
making broad, reliable, and useful inferences based on that material.  Put
simply, training is hard.  Machine learning, deep learning, and
DeepQA~\cite{XYZ0} offer techniques to provide URL.  However, when they are
applied to new domains and new corpuses of information, extensive human
intervention is often required to ensure that the ingested information is
being properly understood, and that reasoning and learning about it is
producing reasonable conclusions.

This human intervention is required in significant measure because the
ingestion process does not have a natural way to assess its efficacy, and
take corrective action when efficacy is found wanting.  Put another way,
human intervention is required to address the problem of unknowns:  AI
ingestion systems do not know what they have not learned.

To address this problem of unknowns, it would help to have an automated means
to generate an interesting and broad set of queries on ingested content, and
that is what we propose here.  To the degree that we can provide good answers
to such automatically generated queries, we know that the ingestion has
worked well.  Further, to the degree we do not provide good answers, the
generated queries provide specific actionable gaps to address.  In turn, the
gaps exposed by specific generated queries may be addressed -- at least in
part -- by looking to other corpuses of data.  In looking at those other
corpuses, we can of course leverage the standard techniques of machine
learning, deep learning, DeepQA, etc. to find the queried information.

This ingest-query combination in turn enables a virtuous cycle:  automated
knowledge extraction -- query generation -- automated knowledge extraction --
etc., where each step checks and expands upon the previous step.  For
example, adding information from additional sources may yield additional
learning based on the cross-product of the original and newly added material.
In some sense this process mirrors the human research endeavor:  understand
material, look elsewhere when key questions are not answered, then pursue new
questions revealed by additional knowledge.

More specifically, this approach leverages three elements:  (1) the automatic
query generation proposed here; (2) DeepQA, machine learning, and deep
learning for resolving the generated queries; and (3) checks on the utility
of the responses from (2).  These three elements in combination provide an
automated means (a) to check if a corpus has been effectively ingested and
understood; (b) to check if the ingested corpus provides broad coverage of
the desired domain; (c) to extend the knowledge in the ingested domain corpus
to cover a broader set of knowledge; and (d) to iterate across (a) - (c) to
continually improve capabilities and accuracy.  In combination, capabilities
(a) - (d) provide an opportunity to automate and scale our cognitive
capabilities.

The remainder of this paper outlines a number of ideas on how to address this
missing link of automatic query generation for ingested corpuses of data.
However, as preview one technique is to apply standard journalistic questions
to the objects (nouns and noun phrases) in an ingested corpus, i.e. {\it who,
what, when, why, where, how.} Applying this technique to an ingested history
corpus might yield, ``{\bf Who} was {\bf General Grant}?''

Of course, automatically generated queries could face the same quality and
understanding problems that face the underlying ingestion process.  To
address this issue and gain a sense of the efficacy of this auto-query
approach, we outline sampling techniques in which humans are used to assess
both individual responses from the automated system, and statically estimate
overall accuracy.

Details are in Section~\ref{section-QueryUtility}.  However, as preview, one
approach to human checking is to leverage crowd-source approaches like Amazon
Mechanical Turk~\cite{XYZ1}.  This approach can provide a measure of how
humans view the quality of the auto-generated queries and added material. In
all of these approaches, human feedback gives a sense of overall accuracy and
helps detect areas where automated knowledge extraction and queries have gone
awry.

\section{Epistemology:  How we know what we know}
\label{section-Epistemology}

The dictionary defines {\it epistemology} as ``a branch of philosophy that
investigates the origin, nature, methods, and limits of human
knowledge.''~\cite{XYZ3} That a branch of philosophy is dedicated to such
investigation and understanding gives some hint of the challenge to
algorithmic understanding of what is known about an ingested corpus.

Following this thought, raises the question, ``What does it mean to
`understand' an input corpus, absent queries about it?''  This inquiry might
be viewed as a reformulation of the proverbial question, ``If a tree falls in
the forest and there is no one there to hear it, does it make a
noise?''~\cite{XYZ4c} However, this inquiry also raises the challenge of
generating meaningful queries about a corpus absent any pre-existing
knowledge of that corpus.  For example, how can one formulate interesting
questions about the US Civil War absent any predefined semantic knowledge
about it?

That challenge is precisely what we propose to address.  To be more specific,
we propose a standardized mechanism for query generation that relies upon
only two things:  (1) the corpus itself in whatever form it has
(traditionally human readable); and (2) well-known structural elements from
the natural language of the corpus, e.g. English.  For example, the system
must know that set of verbs and the set of standard interrogatives.

Combining those two elements we propose a set of six automatic query-generation
techniques outlined next in Sections~\ref{section-query-gen-journalism-objs}
to~\ref{section-query-gen-correlations}.  There are probably additional
techniques for automatic query generation as well.  However, these query
generation techniques can generate a broad set of queries, largely
independent of corpus or domain type.  As such, they suggest a broad set of
approaches that probably suffice to begin, and upon which we now elaborate.

\subsection{Query Generation - Journalism Questions Applied to Objects}
\label{section-query-gen-journalism-objs}

   To every object (noun phrase) in the knowledge base, apply the standard
   questions taught every journalism student:  {\it Who, what, why, when,
   where, how}.

   For example, assume that the knowledge corpus has one statement, ``General
   Grant was in the US Civil War.''  This corpus references two objects:
   {\it General Grand} and {\it US Civil War}.  As a result, using the {\it
   journalism questions} yields two sets of questions, the first about
   General Grant:

   \begin{itemize}
	\item {\bf Who}   was General Grant?
	\item {\bf What}  was General Grant? $\rightarrow$ {\it Prune}
	\item {\bf Why}   was General Grant? $\rightarrow$ {\it Prune}
	\item {\bf When}  was General Grant?
	\item {\bf Where} was General Grant?
	\item {\bf How}   was General Grant?
   \end{itemize}

   The second set of questions is about the US Civil War:

   \begin{itemize}
	\item {\bf Who}   was the US Civil War? $\rightarrow$ {\it Prune}
	\item {\bf What}  was the US Civil War?
	\item {\bf Why}   was the US Civil War?
	\item {\bf When}  was the US Civil War?
	\item {\bf Where} was the US Civil War?
	\item {\bf How}   was the US Civil War?
   \end{itemize}

   Despite their mechanical nature, it can be seen that most of these
   questions are quite reasonable.  Indeed they are the sort of questions
   which if asked in a discussion between two people -- e.g. a teacher and
   student -- would yield not a direct answer, but a great deal of additional
   material, material upon which additional queries may be raised.  For
   example, {\it ``When was General Grant?''} might naturally lead to
   discussion of other events during the years of his life, such as the
   development of Maxwell's Equations in Physics.  Indeed, there are no
   bounds other than sum of human knowledge, and perhaps beyond.

   In automated use, these questions and their follow-ups naturally lead to
   incorporation of material from additional corpuses (to get beyond our
   trival, single-statement corpus of ``General Grant was in the US Civil
   War'').

   Of course, some of the auto-generated questions will be nonsensical, as
   flagged above with {\it Prune}.  In Section~\ref{section-Pruning} we
   discuss some rules and techniques for identifying and pruning such
   nonsensical queries.  Section~\ref{section-QueryUtility} goes further and
   explores how we might determine {\it sensible} queries, that are
   nonetheless not terribly useful.

\subsection{Query Generation - Journalism Questions Applied to Object Pairs}
\label{section-query-gen-journalism-obj-pairs}

   In addition to generating queries on an input corpus by applying the {\it
   journalist questions} to every {\it object}, the journalist questions can
   be applied to every {\it object pair}, e.g.

   \begin{itemize}
	\item {\bf When}:  {\it Was} A after B?

	\item {\bf Where}: {\it Where} is A located relative to B?

	\item {\bf Who}:   $\rightarrow$ ? $\rightarrow$ {\it Prune}

	\item {\bf What}:  $\rightarrow$ ? $\rightarrow$ {\it Prune}

	\item {\bf Why}:   {\it Why} did A \_\_\_ B? 

	   where \_\_\_ is a large set of verbs, and where again pruning is
	   done for nonsensical items as will be described in
	   Section~\ref{section-Pruning}.  Two examples help clarify:

	   \begin{itemize}

	    \item ``{\it Why} did General Grant {\bf fight} the US Civil War?''
	    \item ``{\it Why} did General Grant {\bf eat}   the US Civil War?''
		     $\rightarrow$ {\it Prune}

	   \end{itemize}

	\item {\bf How}:   Similar to {\bf Why}:  {\it How} did A \_\_\_ B? 

	   where \_\_\_ is again a large set of verbs with pruning
           applied to nonsensical verbs, e.g.

	   \begin{itemize}

	    \item ``{\it How} did General Grant {\bf fight} the US Civil War?''
	    \item ``{\it How} did General Grant {\bf eat}   the US Civil War?''
		     $\rightarrow$ {\it Prune}

	   \end{itemize}

   \end{itemize}

\subsection{Query Generation - Comparative Adjectives}
\label{section-query-gen-comparative-adjectives}

   There are additional techniques to automatically generate queries beyond
   the use of {\it journalist questions}.  One such technique leverages {\it
   comparative adjectives} -- an approach that has significant similarities
   to the object-verb-object case just discussed.  For example:

   \begin{itemize}
	\item Is A \_\_\_ B?
   \end{itemize}

   where \_\_\_ is the set of comparative adjectives, e.g.  better, bigger,
   faster, older, closer, hotter, etc.  These {\it compare questions} work
   well in conjunction with {\it object type pairs}, where all obects are
   broken into four types:

   \begin{enumerate}
	\item Person
	\item Object
	\item Location
	\item Concept
   \end{enumerate}

   \begin{table*}

   \begin{center}
   \begin{tabular}{|l|l||l|l|l||l|l|l||l|l|l||}
    \hline
	{\bf Type 1} & {\bf Type 2} &&
	{\bf Type 1} & {\bf Type 2} &&
	{\bf Type 1} & {\bf Type 2} &&
	{\bf Type 1} & {\bf Type 2} \\ \hline \hline
	{\it Person} & {\it Person}  &&      Object  &      Person   &&      Location  &      Person    &&      Concept  &      Person   \\ \hline
	     Person  &      Object   && {\it Object} & {\it Object}  &&      Location  &      Object    &&      Concept  &      Object   \\ \hline
	     Person  &      Location &&      Object  &      Location && {\it Location} & {\it Location} &&      Concept  &      Location \\ \hline
	     Person  &      Concept  &&      Object  &      Concept  &&      Location  &      Concept   && {\it Concept} & {\it Concept} \\ \hline
   \end{tabular}

   \caption{Combinations of Object Pairs}
   \label{Table-Object_Pairs}

   \end{center}

   \end{table*}

   Comparisons are generally meaningful when done between object pairs of the
   same type, i.e. the diagonal elements in Table~\ref{Table-Object_Pairs}.
   Comparisons across object types sometimes make sense as well -- depending
   on the individual verb, as will be described in more detail in
   Section~\ref{section-Pruning}.

\subsection{Query Generation - Analogies}
\label{section-query-gen-analogies}

   The next technique for generating queries has similarities to the
   comparative adjective approach just discussed in
   Section~\ref{section-query-gen-comparative-adjectives}.  That technique is
   to generate queries based on analogies.  For example,

   \vspace{0.1in}

	\hspace{0.5in} {\it ``Who / what} is most {\bf like} {\it object A?''}

   \vspace{0.1in}

   where {\it object A} may be a tree, a horse, the ``French Revolution'',
   etc.

   \vspace{0.1in}

   The answer to this question can also be checked in automated fashion via a
   {\it Reverse check:}  Assume that {\it Object B} was returned in answer
   to the question about what is most like {\it Object A}. Then the following
   question can be posed to the system with the ingested corpus:

   \vspace{0.1in}

	\hspace{0.5in} {\it ``Who / what} is most {\bf like} {\it object B?''}

   \vspace{0.1in}

   Then an obvious check is whether {\it object A} is one of the top
   candidates from this query about {\it object B}.

   There are cases where {\it B} may be most like {\it A}, but the
   relationship is not symmetric, and {\it A} is not most like {\it B}.
   However, the question acts as a filter in terms of what material is
   forwarded to more elaborate checks.

   Those more elaborate -- but still semi-automated checks could include
   things such as the use of {\it Mechanical Turk} on a subset of discovered
   analogies, as also suggested in~\cite{XYZ5}.

\subsection{Query Generation - Extensions to Analogies}
\label{section-query-gen-analogy-extensions}

   Analogies as just described in Section~\ref{section-query-gen-analogies}
   also yield two natural extensions -- questions of the form:

	\begin{enumerate}

	  \item ``Why is that thing / person most like object A?''
	  \item ``What is the evidence and reasoning for that choice?''

	\end{enumerate}

   The second question in particular may require the ingested corpus to use
   DeepQA~\cite{XYZ0} or other techniques to provide appropriate evidence and
   reasoning for the answer to the first question.

\subsection{Query Generation - Correlations}
\label{section-query-gen-correlations}

  Similar to generating queries based on analogies, we can also generate
  questions based on determining what is most strongly correlated with object
  A.  Not all objects are make sense as something on which to perform
  correlations, e.g. a kitchen table.  For purposes of automatic query
  generation, correlations most obviously make sense for objects which are
  generally quantified, e.g. {\it Oil Production}.  However, following the
  discussion in Section~\ref{section-query-gen-comparative-adjectives},
  correlations often make sense for objects of type {\it Concept}, e.g.  {\it
  War} or {\it Popularity}.  For instance, {\it war} may be correlated with
  other concepts such as {\it border dispute} or {\it resource depletion}.

\section{Pruning Nonsensical Queries}
\label{section-Pruning}

Referring back to Section~\ref{section-query-gen-journalism-objs} and the
sample questions about {\it General Grant} and the {\it US Civil War}, recall
that three of the questions about objects in the corpus immediately seemed
silly:

   \begin{itemize}
	\item {\bf What}  was General Grant? $\rightarrow$ {\it Prune}
	\item {\bf Why}   was General Grant? $\rightarrow$ {\it Prune}
	\item {\bf Who}   was the US Civil War? $\rightarrow$ {\it Prune}
   \end{itemize}

Two straight-forward ways to deal with all three of these queries are (a)
simple grammatical rules; and (b) the object type classification described in
Section~\ref{section-query-gen-comparative-adjectives}, i.e. the
classification of objects into {\it Person, Object, Location, Concept}.  Use
of (a) and (b) immediately yield a set of simple rules that can be used to
prune these questions:

   \begin{enumerate}

	\item	The question {\it what} does not fit with object type {\it
		Person}.

	\item   The question {\it who} does not fit with object types {\it
		Object, Location, Concept}.

   	\item   The question {\it why} does not fit with object types {\it
		Person, Location}.

   \end{enumerate}

\begin{table}

  \centering
  \begin{tabular}{|l|l|c|}
    \hline
    {\bf Question} & {\bf Object Type} & {\it Auto Prune?}\\
    \hline \hline
	Who 	& Person   &	       \\ \hline
	Who 	& Object   & {\bf Yes} \\ \hline
	Who 	& Location & {\bf Yes} \\ \hline
	Who 	& Concept  & {\bf Yes} \\ \hline
		&	   &	       \\ \hline
	What	& Person   & {\bf Yes} \\ \hline
	What	& Object   &	       \\ \hline
	What	& Location &	       \\ \hline
	What	& Concept  &	       \\ \hline
		&	   &	       \\ \hline
	When	& Person   &	       \\ \hline
	When	& Object   &	       \\ \hline
	When	& Location &	       \\ \hline
	When	& Concept  &	       \\ \hline
		&	   &	       \\ \hline
	Why	& Person   & {\bf Yes} \\ \hline
	Why	& Object   & {\bf Yes} \\ \hline
	Why	& Location & {\bf Yes} \\ \hline
	Why	& Concept  &	       \\ \hline
		&	   &	       \\ \hline
	Where	& Person   &	       \\ \hline
	Where	& Object   &	       \\ \hline
	Where	& Location &	       \\ \hline
	Where	& Concept  & {\bf Yes} \\ \hline
		&	   &	       \\ \hline
	How	& Person   &	       \\ \hline
	How	& Object   &	       \\ \hline
	How	& Location &	       \\ \hline
	How	& Concept  &	       \\ \hline
   \end{tabular}
\caption{Pruning rules to eliminate nonsensical queries based on specific
combinations of ``journalist questions / interrogatives'' and the four types
into which we group objects.}
\label{Table-Prune-Grammatical1}
\end{table}

More generally, Table~\ref{Table-Prune-Grammatical1} lists a full set of
pruning rules combining the journalist interrogatives and our four object
types.  This pruning based on grammatical structure and object types can be
used beyond the {\it interrogative - object type} pairings in
Table~\ref{Table-Prune-Grammatical1}, and can be extended to the cases in
Section~\ref{section-query-gen-journalism-obj-pairs}, where we have queries
involving questions about object pairs $A$ and $B$ in the corpus and of the
form:

   \begin{itemize}
      \item {\it How} $<$Object A$>$ - $<$Verb$>$ - $<$Object B$>$?
      \item {\it Why} $<$Object A$>$ - $<$Verb$>$ - $<$Object B$>$?
   \end{itemize}

As noted in Section~\ref{section-query-gen-journalism-obj-pairs}, some
queries of this form can be valuable probes, while others are nonsense:

   \begin{itemize}
      \item ``{\it Why} did General Grant {\bf fight} the US Civil War?''
      \item ``{\it Why} did General Grant {\bf eat}   the US Civil War?''
	      $\rightarrow$ {\it Prune}
   \end{itemize}

\begin{table}

  \centering
  \begin{tabular}{|l|r|c|r|c|}
    \hline
    {\bf Q} & {\bf Type of A} & {\bf Verb}
	    & {\bf Type of B} & {\it Auto}\\
    {\bf  } & {\bf          } & {\bf Instance}
	    & {\bf          } & {\it Prune?}\\
    \hline \hline
	Why &   Person A & Verb &   Person B & {\it Prune?} \\ \hline
	Why &   Person A & Verb &   Object B & {\it Prune?} \\ \hline
	Why &   Person A & Verb & Location B & {\it Prune?} \\ \hline
	Why &   Person A & Verb &  Concept B & {\it Prune?} \\ \hline

	Why &   Object A & Verb &   Person B & {\it Prune?} \\ \hline
	Why &   Object A & Verb &   Object B & {\it Prune?} \\ \hline
	Why &   Object A & Verb & Location B & {\it Prune?} \\ \hline
	Why &   Object A & Verb &  Concept B & {\it Prune?} \\ \hline

	Why & Location A & Verb &   Person B & {\it Prune?} \\ \hline
	Why & Location A & Verb &   Object B & {\it Prune?} \\ \hline
	Why & Location A & Verb & Location B & {\it Prune?} \\ \hline
	Why & Location A & Verb &  Concept B & {\it Prune?} \\ \hline

	Why &  Concept A & Verb &   Person B & {\it Prune?} \\ \hline
	Why &  Concept A & Verb &   Object B & {\it Prune?} \\ \hline
	Why &  Concept A & Verb & Location B & {\it Prune?} \\ \hline
	Why &  Concept A & Verb &  Concept B & {\it Prune?} \\ \hline
   \end{tabular}
\caption{Pruning rules to eliminate nonsensical queries based on specific
combinations of ``Why'' and ``How'' along with two objects from a corpus and
a verb connecting those objects.  The verb expands to a long list of verbs
from the language of the corpus.  Note that the form for pruning ``How'' is
completely parallel to ``Why'' and omitted here for brevity.}
\label{Table-Prune-Grammatical2}
\end{table}

To handle pruning of nonsensical questions involving such object pairs, we
take a combined grammatical / semantic approach similar to that for single
objects and which was summarized in Table~\ref{Table-Prune-Grammatical1}.
For object pairs, a much larger table results.  Its structure is sketched in
Table~\ref{Table-Prune-Grammatical2}.  Completing this table with all verbs
from a specific language (e.g. English) requires significant manual effort.
However, that effort is one-time and can then be applied any corpus or topic.

An additional pruning mechanism is to assume any question for which all
responses come back with low confidence~\cite{XYZ0} is a nonsense question.
Such ``low-confidence'' questions could be tracked over time, and tagged as
``nonsense'' historically.  This historical perspective allows determination
of the frequency with which questions change over time from ``nonsense'' to
``non-nonsense'' -- because with new knowledge, a high-confidence response
emerges to a question previously assessed as nonsense due to low-confidence
responses.  The degree to which such classification changes occur gives an
estimate of how many questions are incorrectly marked nonsense historically.

Even with this historical perspective, this method of pruning by confidence
could result in missing key gaps in the corpuses and knowledge bases.  Thus,
the pruning rules outlined in Tables~\ref{Table-Prune-Grammatical1}
and~\ref{Table-Prune-Grammatical2} may be a better initial approach, coupled
with experimentation around confidence-based categorizations of questions.

\section{Utility of Automatically Generated Queries}
\label{section-QueryUtility}

Even after pruning as described in Section~\ref{section-Pruning}, the
techniques described in Section~\ref{section-Epistemology} naturally generate
a very large set of queries on an ingested corpus.  Beyond the need to form
sensible queries as outlined in Section~\ref{section-Pruning}, it is also
important to get a practical sense of the value of the queries in
understanding the corpus.  Some sensible queries may still yield ``boring''
results.  For example, using queries with comparative adjectives as described
in Section~\ref{section-query-gen-comparative-adjectives}, we might generate
the query, {\it ``Is a paper clip taller than a building?''}  The answer to
this query could be useful in some analyses, but the query itself is unlikely
to be asked by a human.

Of course, questions that may superficially or initially seem silly can
reveal deep, interesting responses.  For example, consider the auto-generated
query, {\it ``Is lithium older than iron?''} The answer turns out to be true,
based on the development of the universe since Big Bang as described by Hoyle
and others in the theory of Stellar Nucleosynthesis~\cite{XYZ6}.

As both a theoretical and practical matter, it is helpful to know how many
auto-generated queries are useful.  In terms of theoretical value, knowing
how many and which queries and query types yield useful responses is
important in the development of further rules and heuristics to determine
{\it a priori} which auto-generated queries make sense.  In terms of
utilitarian value, having insight into how to prune is important in terms of
keeping computation and storage tractable.  Knowing the faction of useful
queries is also important for some of the techniques for checking results and
filling gaps, as will be discussed in Section~\ref{section-FillingGaps}.

Section~\ref{section-Pruning} suggested using confidence of responses to an
auto-generated query as a measure of the query's sensibility.  This
confidence measure can also be used as a measure of a query's utility.
Indeed, it is both an advantage and a drawback of this confidence method that
for low confidence results, it may be difficult to disambiguate whether the
cause was a nonsensical query or a boring query.  (For high confidence
results, the query was presumably both sensible and interesting.)  Further
research is needed to understand these distinctions and their underpinnings.

As outlined in the Introduction, approaches like Mechanical Turk~\cite{XYZ1}
or Crowdflower~\cite{XYZ2} can also be used to check both sensibility and
utility on a sample of auto-generated queries and their responses.

Another ``human'' oriented approach to checking the utility of auto-generated
queries is to update or create Wikipedia entries when key missing information
is detected, and where the cognitive system infers (with high confidence)
that it has information to fill the gap.  If these cognitive updates and
entries pass muster with Wikipedia editors, then they are good.  The
cognitive system could even check periodically {\it (a)} if previously
proposed updates have been maintained (i.e. the updated info was correct and
useful); or {\it (b)} if previously proposed updates have been removed, i.e.
the info was incorrect or unhelpful.

Both of these ``human'' approaches can yield statistical information about
the fraction of auto-generated queries that are useful, as well as specific
examples of such queries that are useful and not.  However, these ``human''
approaches cannot broadly label {\em all} auto-generated queries into the
desired three categories:  (1) useful and interesting; (2) useful, but not
interesting; and (3) nonsensical.

\section{Filling Gaps Revealed by Auto Queries}
\label{section-FillingGaps}

As we have noted in several times previously, it is important to have some
way to complete the cycle and fill-in incomplete data revealed by
auto-generated queries.  For two practical reasons, it is probably not
feasible to auto-generate queries across all ingested corpuses of
information:

\begin{enumerate}

  \item Two of the automatic query generation techniques (i.e. the techniques
	in Sections~\ref{section-query-gen-journalism-obj-pairs}
	and~\ref{section-query-gen-comparative-adjectives}) yield a number of
	queries that is quadratic in the number of total objects across all
	corpuses.  Looking just at words in English, a quadratic function
	could yield a trillion object pairs, as~\cite{XYZ8} reported over 1
	million English words.  And not only are there many languages, but
	objects can be noun phrases and other complex arrangements.

  \item Even if it is computationally tractable to handle the number of
	object pairs in (1), increased noise seems likely to result from
	considering all corpuses simultaneously when generating queries.  For
	example, generating queries based on object pairs from a corpus on
	medieval poetry and a corpus on protein folding is unlikely to yield
	queries of high utility of the type outlined in
	Section~\ref{section-QueryUtility}.  Aside from the computational
	problems noted in (1), having a high percentage of low-utility
	queries is likely to be a problem when humans evaluate query utility
	as also described in~\ref{section-QueryUtility}.  People frustrated
	by a high percentage of silly material are likely to provide lower
	quality checking than people reviewing results in which silly
	material is less common.

\end{enumerate}

If practical considerations militate against generating queries from every
pair of corpuses simultaneously, there is still the question of which sets of
corpuses should be paired.  A straight-forward (if somewhat dumb and
unsophisticated) mechanism for making this determination is {\it (a)} to
generate queries from every pair of corpuses and then {\it (b)} use measures
of sensibility and utility from Sections~\ref{section-Pruning}
and~\ref{section-QueryUtility} on the generated queries to determine which
corpuses to pair.  In other words, pair those corpuses that yield the highest
percentage of useful queries.

Transitive closure could then be used to combine corpus pairs into corpus
groupings with cardinality greater than 2.  Such closure could stop when the
percentage of useful queries between a pair of corpuses falls below a
threshold.  Of course, techniques other than query generation could be used
in forming corpus groupings, as has been explored elsewhere~\cite{XYZ0}.

Note also that the set of corpuses paired for generating queries need not be
the same as the set of corpuses examined in fielding these queries.  Indexing
and tagging individual corpuses~\cite{XYZ10} may provide criteria by which
corpuses are selected to resolve auto-generated queries.  

A further wrinkle to the question of corpus merging arises from the fact that
the corpuses themselves can change as a result of the query-generation
process.  More specifically, the pairing of auto-query generation and
fielding those queries can create new enlarged corpuses in which particular
pieces of information and their connections are explicitly noted -- as
opposed to requiring inference.  At one level, this corpus merging and
enlargement and explicit notation of connections could be considered ``mere
optimization''.  However, this optimization is akin to humans learning a new
piece of information and not needing to re-derive it every time it is needed.
Such capability has clearly provided great benefit in the biological
realm~\cite{XYZ11}, e.g. as a means to reduce energy requirements for an
organism, and thus reduce food needs.  In biology this learning capability
can also provide faster response time, thus allowing escape from a dangerous
situation that would not otherwise be possible.  As such, it seems plausible
that such optimization will have similar value here.

As noted in the Introduction, this amalgamation of corpuses and
corresponding learning process also mirrors the human research endeavor:
understand material, look elsewhere when key questions are not answered,
then pursue new questions revealed by additional knowledge.

As with human research and the scientific method~\cite{XYZ12}, another
component of completing this cycle of query generation and fielding those
queries is to maintain historical confidence levels for inferences.  Then
periodically look for new information that would increase or decrease
confidence in those inferences.  

The most effective combination of techniques and heuristics to use for
completing this cycle of query generation and query response is a subject
of further research.

\section{Measuring Capability and Changes to It}
\label{section-Measurement}

With all of the capabilities described in the previous sections, it
would be useful to have simple, high-level measures of overall
capability and changes to it.  Following the lead of~\cite{XYZ0}, we
focus on a pair of measures:  {\it (a)} breadth of coverage and {\it
(b)} precision.  {\it (a)} Breadth of coverage is the percentage is
auto-generated queries for which the system can provide a
high-confidence reply.  {\it (b)} Precision measures the percentage of
all attempted responses in {\it (a)} which are correct.

Unlike IBM's Watson playing Jeopardy~\cite{XYZ0}, precision itself is a
fuzzy measure in this realm, since there is no answer book which says
for all queries whether the given response is correct.  Thus precision
must be determined by some combination of other means.  Thankfully, we
do not need an exact measure of precision to get a general sense of how
the system is performing, and the degree to which it is improving or
degrading over time.

Human-based techniques described in Section~\ref{section-QueryUtility}
and leveraging Mechanical Turk, Wikipedia, etc provide one set of
baselines -- with Mechanical Turk perhaps giving a sense of average
human ability in a particular domain and Wikipedia giving a sense of
more expert human ability.  For topics of special importance, domain
experts could be recruited or contests held, a la Jeopardy -- perhaps
handicapping the automated system by limiting it to a ``small'' amount
of training time.

All of these approaches allow some answer to key questions about our
proposed system:  {\it (1)} can it answer as many questions as a human
or more, {\it (2)} can it do it with equal or greater precision, and
{\it (3)} can it do so in equal or shorter time?

External benchmark questions could also be used for these measurements
such as from the Stanford QA DB~\cite{XYZ13}.  Such benchmark questions
should of course be applied after our proposed system has trained itself
on auto-generated queries and an enlarging set of corpus material, as
outlined in Section~\ref{section-FillingGaps}.

\section{Related Work}
\label{section-RelWork}

Previous work on automatic query generation has largely focused on
specialized domains.  For example, \cite{XYZ7b}, ``generate web-search
queries for collecting documents matching a minority concept.''  
\cite{XYZ7c} explore, ``interactive construction of natural language
queries.''  \cite{XYZ7d} process patent applications looking for terms that
can be used to generate queries automatically.  \cite{XYZ7e} automatically
generate queries based on ``keywords expanded from [a] user input keyword,
$\ldots$ by selecting candidate keywords and assigning weight value[s] to
each candidate keyword from [a] semantic thesaurus and document keyword
list.''  Unlike the approach described here, the approach of Ryu et al
requires some domain knowledge on the part of the user.  None of these other
works address automatic generation of queries for a general corpus on
arbitrary topics and in arbitrary format.

There is a good deal of work that uses grammatical structure for pruning as
discussed in Sections~\ref{section-Pruning} and~\ref{section-QueryUtility}.
For example, \cite{XYZ8b} prune noun phrases.  \cite{XYZ8c}, ``present a
hybrid method for text summarization, combining sentence extraction and
syntactic pruning.''  However, these works focus primarily on identification
of particular grammatical artifacts and reduction of corpus size, as opposed
a more semantic view of comprehending content, as proposed here.  Broader
measures of confidence and pruning were well explored in~\cite{XYZ0}.

The epistemology approach described in Section~\ref{section-Epistemology} is
most closely related to~\cite{XYZ9}.  That paper, ``studies what kinds of
facts about the world are available to an observer with given opportunities
to observe, how these facts can be represented in the memory of a computer,
and what rules permit legitimate conclusions to be drawn from these facts.''
As such it focuses more on what is knowable versus the focus here on
understanding and expanding what has been ingested.

The techniques described here can be viewed as a combination of supervised
and unsupervised learning techniques.  In particular, the cycle of {\it
auto-query generation / find missing material / auto-query generation on
expanded corpus / etc} is a form of unsupervised learning, which dates
to~\cite{XYZ14}.  Unlike that work and its successors, the approach here is
not based on a neural nets or statistical approaches, but a variety of
grammatical and semantic techniques applicable to unstructured content such
as text.  Techniques described in Section~\ref{section-QueryUtility}, where
results are reviewed by humans, represent an offline form of supervised
learning.  Like unsupervised approaches, much of the existing supervised
literature focuses on neural nets, as typified by~\cite{XYZ15} on
back-propagating errors.

Aside from supervised and unsupervised learning, most learning applications
focus on specific domains, e.g.~\cite{XYZ16,XYZ17} or learning
representations, e.g.~\cite{XYZ18,XYZ19}.  Our techniques are cross-domain,
and do not focus or require particular representations.

\cite{XYZ20} discuss learning by crowd sourcing, which has similarities to
techniques in Section~\ref{section-QueryUtility} using Mechanical Turk.
\cite{XYZ21,XYZ22} have proposed a set of techniques to write Wikipedia
articles automatically.  However, their approach is based on finding similar
articles, abstracting and summarizing them, and then using that
representation to generate Wikipedia articles, or additions to existing
articles.  By contrast, our approach looks for missing material (via the
auto-generated queries) and adds it to Wikipedia.

\section{Conclusions}
\label{section-Conclusion}

We have described techniques to generate queries automatically based on one
or more ingested, input corpuses.  These queries can then be used to test the
efficacy of ingestion -- how much does the system really know.  Since the
queries are automatically generated, neither their generation nor testing of
ingestion based on the queries requires human experts.

By removing the human factor, this approach allows more effective scaling of
AI and cognitive capabilities and more rapid deployment to new domain areas.
By means of the generated queries, this approach also provides a means to
identify and remedy problem areas in ingestion.
The proposed techniques also allow knowledge to be extended naturally, and
for tests of capability.

Proving the approaches outlined here is, of course, essential, and we look
forward to reporting the results of such investigations.

\end{document}